\documentclass[10pt, conference]{IEEEtran}
\IEEEoverridecommandlockouts

\usepackage{cite}
\usepackage{amsmath,amssymb,amsfonts}

\usepackage{graphicx}

\usepackage{xcolor}
\usepackage{xurl}
\usepackage[hidelinks]{hyperref}
\usepackage{siunitx}
\usepackage{booktabs}
\usepackage{threeparttable}

\usepackage[shortlabels,inline]{enumitem}
\setenumerate[1]{leftmargin=1.0cm}

\newcommand{\sus}{\textsc{Console}}

\begin{document}

\bstctlcite{IEEEexample:BSTcontrol}

\title{Anomaly Detection in Large-Scale Cloud Systems: An Industry Case and Dataset}

\author{
    \IEEEauthorblockN{Mohammad Saiful Islam\IEEEauthorrefmark{1}, Mohamed Sami Rakha\IEEEauthorrefmark{1}, William Pourmajidi\IEEEauthorrefmark{1}, Janakan Sivaloganathan\IEEEauthorrefmark{1}, \\ John Steinbacher\IEEEauthorrefmark{2}, and Andriy Miranskyy\IEEEauthorrefmark{1}}
    \IEEEauthorblockA{\IEEEauthorrefmark{1}Dept. of Computer Science, Toronto Metropolitan University, Toronto, Canada, \\ Email: \{mohammad.s.islam, rakha, william.pourmajidi, jsiva, avm\}@torontomu.ca}
    \IEEEauthorblockA{\IEEEauthorrefmark{2}Cloud Platform, IBM Canada Lab, Toronto, Email: jstein@ca.ibm.com}
}

\maketitle
\thispagestyle{plain}
\pagestyle{plain}

\begin{abstract}
As Large-Scale Cloud Systems (LCS) become increasingly complex, effective anomaly detection is critical for ensuring system reliability and performance. However, there is a shortage of large-scale, real-world datasets available for benchmarking anomaly detection methods.

To address this gap, we introduce a new high-dimensional dataset from IBM Cloud, collected over 4.5 months from the IBM Cloud Console. This dataset comprises 39,365 rows and 117,448 columns of telemetry data. Additionally, we demonstrate the application of machine learning models for anomaly detection and discuss the key challenges faced in this process.

This study and the accompanying dataset provide a resource for researchers and practitioners in cloud system monitoring. It facilitates more efficient testing of anomaly detection methods in real-world data, helping to advance the development of robust solutions to maintain the health and performance of large-scale cloud infrastructures.

\end{abstract}

\section{Introduction}

In recent decades, the adoption of Cloud Computing across government and business sectors has grown exponentially~\cite{buyya2013mastering, gartner2024cloud}. At the core of this growth is the ability of cloud computing to offer high-capacity data centers as a reliable backbone for services. Cloud providers operate expansive data centers that support global workloads, necessitating sophisticated techniques to monitor, diagnose, and respond to failures in real-time. As cloud infrastructure expands in both scale and complexity, maintaining its reliability has become a critical concern. Even brief outages or performance issues can lead to significant losses for users hosting applications in the cloud~\cite{chkirbene2020machine}.

To prevent such issues, cloud system administrators must continuously monitor hardware and software services to ensure compliance with service-level agreements (SLAs)~\cite{wu2015service}. System anomalies, which translate into unexpected behavior, reduced efficiency, or even downtime, pose a significant risk. Early detection of these anomalies is vital for taking preemptive measures to safeguard users, improve the overall user experience, and ensure SLAs. Various studies have introduced anomaly detection methods based on statistical and machine learning techniques, spanning supervised~\cite{farshchi2015anomaly}, semi-supervised~\cite{deka2022semi,gao2021connet}, and unsupervised approaches~\cite{baek2017unsupervised,islam2020anomaly,islam2021anomaly,zhong2024detecting}. These models have been tested using diverse datasets and systems of varying complexity and scale~\cite{soldani2022anomaly,hagemann2020systematic}.

One major challenge for anomaly detection methods is the high dimensionality of data generated in large-scale cloud computing environments~\cite{thudumu2020comprehensive_article}. Many existing methods struggle to maintain accuracy in the presence of this ``curse of dimensionality''~\cite{bellman1961adaptive}, which hampers both performance and precision. High-dimensional data requires more input for generalization and results in data sparsity, where data points become scattered and isolated. The abundance of irrelevant features often obscures true anomalies, reducing the effectiveness of traditional methods such as distance or clustering-based techniques~\cite{cluster_outlier_detection_aggarwal}.

Additionally, much of the existing work in anomaly detection has been conducted on relatively small datasets~\cite{lavin2015evaluating, MicrosoftCloudDataset:online,panahandeh2024serviceanomaly}, which may not fully capture the challenges posed by larger-scale cloud systems. To help advance this area of research, we aim to share a large-scale dataset from a real-world IBM Cloud System~\cite{islam2020anomaly} with the broader community. This will enable more comprehensive testing and evaluation of anomaly detection methods on large, complex datasets.
We address the following research questions (RQs):

\textbf{RQ1:} What are the key characteristics of telemetry datasets collected from Large-Scale Cloud Systems\footnote{Comprising numerous hardware and software components, which are often distributed across multiple data centers.} (LCS)?

\textbf{RQ2:} What are the main challenges in predicting anomalies within such large datasets?

The main contributions of this paper are:
\begin{itemize}
    \item Introducing a new large-scale dataset for testing anomaly detectors in cloud systems. The dataset is available on Zenodo~\cite{islam_2024_14062900}.
    
    \item Demonstrating predictive models for anomaly detection in cloud environments. The reproducibility package is accessible on GitHub~\cite{repro_github} and Zenodo\cite{repro_zenodo}.
    \item Discussing challenges related to handling high-dimensional telemetry data using domain knowledge and machine learning techniques.
\end{itemize}

The remainder of the paper is organized as follows. Section~\ref{sec:lit_review} reviews the related literature. Section~\ref{sec:dataset} introduces the dataset. Section~\ref{sec:anomaly_detectors} details the construction of anomaly detectors. Section~\ref{sec:challenges} discusses the challenges faced. Finally, Section~\ref{sec:conclusion} concludes the paper.

\section{Related Work}\label{sec:lit_review}

\subsection{Existing Datasets and Benchmarks}\label{sec:existing_datasets}
Listed below are popular datasets and benchmarks for detecting anomalies in Cloud systems.

The \textit{NAB (Numenta Anomaly Benchmark) dataset}~\cite{ahmad2017unsupervised, lavin2015evaluating} comprises 57 one-dimensional time series collected from diverse sources such as web traffic, power consumption, and sensor readings~\cite{NABDataset:online}. It includes both real-world and synthetic data intended for testing anomaly detection in real-time streaming applications. Each dataset contains an average of \num{6303} rows, ranging from \num{1127} to \num{22695} rows. While NAB is recognized for its utility in real-time anomaly detection, it has technical weaknesses such as missing values and varying data distributions, limiting its practical utility for high-dimensional anomaly detection tasks~\cite{singh2017demystifying}.

The \textit{Microsoft Cloud Monitoring dataset} consists of 67 one-dimensional time series from production telemetry signals~\cite{MicrosoftCloudDataset:online}. Each dataset includes a timestamp, metric value, and anomaly label fields, capturing metrics like database query rates, service API latency, and application crash rates at per-minute or per-hour intervals. The datasets average \num{3757} rows, ranging from \num{176} to \num{20160} rows. Although useful for evaluating anomaly detection algorithms, its small scale and low dimensionality limit its suitability for modeling complex relationships in high-dimensional cloud telemetry environments.

The \textit{Exathlon dataset} is a high-dimensional dataset constructed from real data traces of repeated executions of large-scale stream processing jobs on an Apache Spark cluster over 2.5 months~\cite{exathlon2021}. It includes 93 traces (after pruning) from 100 executions of 10 distributed streaming jobs, with six types of intentionally introduced anomalies like misbehaving inputs, resource contention, and process failures. Each trace consists of \num{2283} metrics recorded every second for about seven hours on average. While Exathlon serves as a benchmark for explainable anomaly detection in high-dimensional time series and is larger in scale and dimensionality than NAB, it focuses on a fixed set of repeated streaming tasks, which may limit its applicability for capturing the complexities and variability of dynamic cloud environments.

We will compare these datasets with ours in Section~\ref{sec:dataset_comparison}.

\subsection{Anomaly Detection Methods}
We categorize the anomaly detection methods as follows.

\paragraph{Supervised Methods} Supervised approaches treat the anomaly detection problem as binary classification. With complete and accurate ground truth labels, supervised classifiers can detect known anomalies but may miss unknown ones. Typically, existing classifiers like Neural Networks~\cite{lecun2015deep} and Random Forest~\cite{breiman2001random} are employed. A key challenge is that ground truth labels may not cover the full spectrum of anomaly types, limiting the ability of supervised methods to identify unfamiliar or unlabeled anomaly patterns~\cite{chalapathy2019deep}.

\paragraph{Semi-Supervised Methods} Semi-supervised anomaly detection algorithms leverage partially available labels while retaining the capability to detect unseen anomalies. Recent studies use partially labeled data to enhance detection accuracy and exploit unlabeled data for representation learning. Some semi-supervised models are trained exclusively on normal samples, identifying anomalies that deviate from learned normal representations~\cite{ruff2019deep, chalapathy2019deep}.

\begin{figure*}[t]
    \vspace*{-6mm}
    \includegraphics[width=1.0\textwidth]{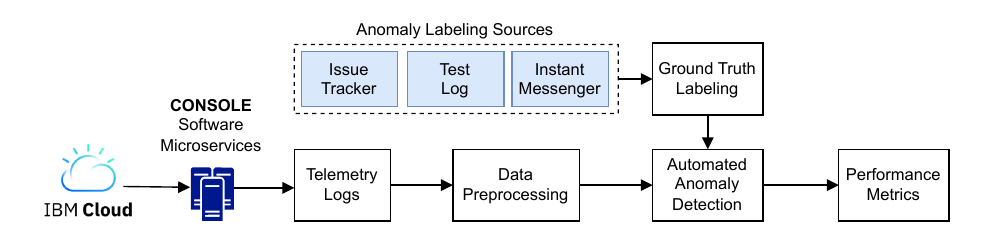}
    \caption{Overview of the data pipeline.}
    \label{fig:Overview}
\end{figure*}

\paragraph{Unsupervised Methods} Unsupervised anomaly detection methods operate under various assumptions about data distribution~\cite{hodge2004survey}, such as anomalies residing in low-density regions. Their performance depends on how well the input data aligns with these assumptions. Numerous unsupervised methods have been proposed~\cite{chandola2009anomaly}, broadly categorized into shallow and deep (neural network) methods. Shallow methods often offer greater interpretability, while deep learning (DL) methods excel with large, high-dimensional data. Unsupervised DL methods for anomaly detection in multivariate time series have gained significant attention due to the increasing complexity and dimensionality of software systems monitoring.

Unsupervised DL methods can be categorized along three key dimensions: (1) inter-variable correlation evaluation~\cite{zhang2019deep}, (2) temporal context modeling~\cite{hundman2018detecting}, and (3) anomaly score criteria~\cite{zong2018deep}. The first dimension involves methods for quantifying correlations among multiple variables, such as dimensional reduction, 2D matrices, or graphs~\cite{lee2007nonlinear}, allowing high-dimensional monitoring data to be succinctly represented with reduced feature sets, mitigating dimensionality issues and computational resource requirements. The second dimension focuses on the temporal context within time series data, depending on the choice of neural network architectures like Recurrent Neural Networks (RNNs)~\cite{rumelhart1986learning}, Long Short-Term Memory (LSTM)~\cite{hochreiter1997long}, Gated Recurrent Units (GRUs)~\cite{cho2014learning}, or Convolutional Neural Networks (CNNs)~\cite{lecun1998gradient}. The third dimension revolves around determining anomaly scores that indicate levels of anomalousness; higher scores suggest a greater likelihood of abnormal behavior. Anomaly score calculations include methods like reconstruction error~\cite{sakurada2014anomaly}, prediction error~\cite{hundman2018detecting}, or dissimilarity metrics~\cite{ruff2020unifying}.

In our prior work~\cite{islam2020anomaly,islam2021anomaly}, we proposed an anomaly detection architecture based on a GRU-based autoencoder with a likelihood function for anomaly detection in multi-dimensional cloud telemetry. Our initial results showed that this model could detect anomalies up to 20 minutes earlier than previous monitoring systems and significantly reduced false alerts. The detector's performance was validated against both publicly available benchmark datasets and real-world data from the IBM Cloud Platform. 
That work focused on an earlier version of the software under study, which has significantly evolved over the past few years, becoming more sophisticated yet increasingly complex (in line with the Laws of Software Evolution~\cite{lehman1996laws}). 
Since 2021, the IBM Cloud Platform has undergone significant changes, including migration to containerized environments and the implementation of new technologies, leading to data with new characteristics that require tailored anomaly detection methods to address evolving challenges effectively. 

In this paper, we aim to highlight the challenges raised by the IBM Cloud Platform's evolved system, particularly in working with the new data it generates. 
Unlike the previous studies~\cite{islam2020anomaly,islam2021anomaly}, which did not share the data or models, this work provides both, enabling the community and industry to adopt the dataset as a benchmark for detection of anomalies in operation of large-scale anomaly cloud software. Additionally, we examine the difficulties in identifying ground truth labels and constructing automatic anomaly detection techniques on these complex datasets.

\section{Dataset Creation and Description}\label{sec:dataset}
\subsection{Software System Under Study}\label{sec:sus}
In this study, we collected data from the IBM Cloud Console (hereon referred to as the \sus{}), the primary web interface and orchestrator\footnote{The \sus{} supports key functions like identity management, billing, search, tagging, and access to the product catalog.} for IBM Cloud~\cite{islam2021anomaly}. IBM Cloud is a public cloud infrastructure with a global network of over 60 data centers~\cite{IBMCloud84:online}. Figure~\ref{fig:Overview} presents an overview of the data pipeline in this study, illustrating the process from data extraction, through model training, to performance evaluation.

The \sus{} production software system is deployed across seven data centers worldwide, though not all instances are active at the same time; they are rotated based on operational requirements. Additional \sus{} deployments are spread globally for testing and staging environments. The \sus{} uses a microservices architecture, with each microservice generating millions of daily logs and telemetry records (a common challenge with large scale Cloud systems~\cite{miranskyy2016operational,pourmajidi2017challenges,pourmajidi2019dogfooding,pourmajidi2021challenging}). These records offer valuable insights into the health and performance of the IBM Cloud, providing critical data for monitoring and optimization. The sheer volume and variability of logs and telemetry, combined with the dynamic nature of containerized environments~---~characterized by transient workloads, volatile resource usage, and continuous scaling~---~necessitate advanced, context-aware anomaly detection techniques. Traditional static or threshold-based methods fall short in addressing the complexity and scale of these systems. Our anomaly detection approach aims to address these challenges by leveraging innovative techniques designed for dynamic environments, contributing to more effective monitoring and actionable insights.

\subsection{Data Collection Methodology}
 
In this paper, we collected telemetry data from the IBM Cloud \sus{}, including logs and metrics generated by its microservices. Due to the large volume of logs and telemetry emitted by the microservices, we used a Publish/Subscribe (Pub/Sub) mechanism~\cite{birman1987exploiting} to efficiently manage the data collection. The microservices publish their logs through a Redis Pub/Sub system~\cite{RedisMes98:online} as part of this process.

We developed a resilient pipeline for the collection and analysis of real-time logs from Redis Pub/Sub, using our previous research~\cite{hoque2018architecture, hoque2018online, islam2021anomaly}. This pipeline is data-agnostic, capable of receiving, processing, and storing data in the IBM Cloud Object Store (COS)~\cite{IBMCloud33:online} for near-real-time analysis.

A sub-pipeline called ``Firehose'' subscribes to Redis Pub/Sub, continuously receiving log data in 
Zipkin format\footnote{At the time of writing, this sub-pipeline was upgraded to receive data in Open Telemetry format~\cite{opentelemetry}.}~\cite{DataMode69:online}. Our containerized microservices, managed by Kubernetes, connect to Firehose to perform Extract, Transform, and Load operations, after which the processed data are stored in IBM COS. These microservices are managed by IBM DevOps tools~\cite{DevOpsSo81:online} and toolchains~\cite{Usingtoo99:online} within a Continuous Integration/Continuous Delivery framework.

In this study, we collected a large dataset over $\approx$ 4.5 months~---~from January~22, 2024, to June 7,~2024. The data comes from seven production data centers during this period, providing a comprehensive view of the \sus{} system’s performance.

\subsection{Dataset Description}\label{sec:dataset_description}
The collected \sus{} dataset provides response time information for individual requests processed by software microservices, aggregated over 5-minute intervals. The choice of a 5-minute interval was based on feedback from the IBM Operations (Ops) team to reduce noise and improve data usability. The response time aggregation was performed using eight statistical functions: minimum, maximum, median, average, count, standard deviation, skewness, and kurtosis.

The final tabular dataset contains a total of \num{39365} rows, each row representing a 5-minute interval. The dataset includes one column for the start time of each interval and \num{117448} columns for the aggregated statistics. The column names specify details such as the datacenter, host microservice endpoint, request type, and response code. Details on data preprocessing are provided in Appendix~\ref{sec:data_preprocessing}. The characteristics of the dataset are provided in Appendix~\ref{sec:data_characteristics}. 

\subsection{Annotation Process for Anomalies}\label{sec:anomaly_annotation}
\subsubsection{Anomaly Labelling Sources}
We created ground-truth labels for anomalies using data from three sources, including an issue-tracking system, test log monitoring, and an internal instant messaging platform~\cite{ligus2013effective} (see Figure~\ref{fig:Overview}). Each source offered a distinct perspective on system anomalies, enabling us to construct a comprehensive and accurate representation of the software system issues.

\paragraph{Issue Tracker}
This tracker focuses on customer-impacting events (such as disruptions affecting customer experience, service access, or quality). Typically, these issues were followed by root cause analysis, which provide additional insights into their origins.

\paragraph{Test Log}
This tracker recorded alerts triggered by failures in synthetic UI test cases or heartbeat signal disruptions. Incidents were logged when failure thresholds were exceeded to minimize false alarms, signaling potential system issues.

\paragraph{Instant Messenger}
We also monitored IBM Ops instant messaging communications, where potential anomalies were informally discussed. Only incidents confirmed by the IBM Ops team were included in the ground truth labels.

By integrating data from these three sources, we ensured that the ground-truth labels represented a comprehensive view of system anomalies. \textit{In total, we identified 3 anomalies from the Issue Tracker, 11 anomalies from the Test Log, and 11 anomalies from the Instant Messenger.}

\subsubsection{Anomaly Timing}
The start and end times of anomalies were estimated based on expert guidance from the IBM Ops team. The start time was set 20 minutes before the issue was reported, reflecting the typical delay between detection, initial triage, and the creation of a ticket or instant message. End times were derived from duration data recorded in the Issue Tracker, Testing Log, or Instant Messenger based on human confirmation of issue resolution.

From our discussion with the IBM Ops team, we learned that some anomalies were mitigated by measures such as redirecting traffic to healthy instances as a high availability strategy. This approach often allowed anomalies to persist on certain DC instances for extended periods (e.g., hours) without noticeably affecting the user experience. However, the IBM Ops team was still responsible for identifying and resolving the underlying root causes of these anomalies to ensure long-term system stability.

\subsection{Comparison with Existing Datasets}\label{sec:dataset_comparison}
The existing benchmark datasets (as discussed in Section~\ref{sec:existing_datasets}) are widely used for comparing anomaly detection models in cloud environments. However, they often have limitations such as low dimensionality, with feature counts ranging from \num{1} to \num{2283}, reliance on synthetic data, and a focus on static tasks (for Exathlon). These constraints reduce their relevance for the dynamic and large-scale nature of real-world cloud systems.

In contrast, our dataset is distinguished by its use of live data\footnote{The \sus{} microservices handled $\approx$ \num{3.2E+09} requests during the 4.5~month period (based on the sum of all aggregated\_stats\_value fields where aggregated\_stats\_name is equal to \texttt{count}).}, extended monitoring period ($\approx$ 4.5 months), high dimensionality ($\approx$ \num{1.1e5}), and focus on a real cloud environment (from IBM Cloud). These factors make it more suitable for analyzing and predicting anomalies in large-scale cloud-based systems, where capturing interactions between various components over time is crucial in large systems. The key statistics of the datasets are summarized in  Table~\ref{tbl:datasets}.

\begin{table}[tb]
    \caption{Comparison of Datasets for Anomaly Detection in Cloud Systems.}
    \label{tbl:datasets}
    \centering
    \begin{tabular}{@{}lrrrr@{}}
        \toprule
        Dataset & File & Column & Avg. Row & Row \\ 
        Name & Count & Count & Count & Range \\
        \midrule
        NAB & \num{57} & \num{1} & \num{6303} & \num{1127}--\num{22695} \\ 
        MS Cloud Monitoring & \num{67} & \num{1} & \num{3757} & \num{176}--\num{20160} \\ 
        Exathlon & \num{93} & \num{2283} & \num{25200} & \num{25200} \\ 
        \textit{Ours} & \num{1} & \num{117449} & \num{39365} & \num{39365} \\ 
        \bottomrule
    \end{tabular}
\end{table}

\section{An Example of Detecting Anomalies}\label{sec:anomaly_detectors}
\subsection{Predictive Models}

To demonstrate how to use the collected dataset for anomaly prediction and model building, we present two simple autoencoders: one based on an artificial neural network (ANN) using Multi-Layer Perceptrons~\cite{Goodfellow-et-al-2016}, and another based on the GRU~\cite{cho2014learning} architecture. The former is simpler, while the latter is more sophisticated. Both models are implemented using the TensorFlow package v.2.15.0~\cite{tensorflow2015-whitepaper}. The architectures of the autoencoders are as follows.

\subsubsection{ANN Autoencoder}
The ANN Autoencoder is a fully connected model well suited for feature-based anomaly detection. It starts with an input layer of \num{2410} dimensions, followed by a series of dense layers that gradually compress the data down to a latent space of 14 neurons. The encoding path reduces the dimensionality with layers of 128 and 64 neurons, each followed by Leaky ReLU activations~\cite{maas2013rectifier}, batch normalization, and dropout for regularization. The decoder then mirrors the encoding structure, expanding back to \num{2410} dimensions. The model contains $\approx$~\num{6.4E+05} trainable parameters.

\subsubsection{GRU Autoencoder}
The GRU Autoencoder is generally designed for sequential data. It processes our \num{2410} featured sequential data, with seven stacked GRU layers compressing the input into a latent space of 14 features. Each GRU layer has 16 units with ReLU~\cite{nair2010rectified} as an activation function, which captures temporal dependencies in the data. After encoding, the latent features are expanded through a repeat vector layer, allowing the decoder to reconstruct the original sequence. The decoder uses seven GRU layers to gradually reconstruct the sequence back to its original length. The model contains $\approx$~\num{1.8E+07} trainable parameters.

\subsubsection{Anomaly Likelihood Function}

For both models, the reconstruction error is passed to the anomaly likelihood function introduced in~\cite{lavin2015evaluating, ahmad2017unsupervised}. The likelihood function is constructed as follows. It maintains a window of the last $W$ error values and processes raw errors incrementally. Historical errors are modeled as a rolling normal distribution of a window of the last $W$ points at each step $t$. The empirical mean $\mu_t$ and standard deviation $\sigma_t$ at time $t$ are computed as follows:
\begin{equation}\label{eq:mu}
    \mu_t = \frac{\sum_{i=0}^{W-1}{s_{t-i}}}{W},
\end{equation}
where $s_{(\cdot)}$ is the prediction error computed by the model, and
\begin{equation}\label{eq:sigma}
    \sigma_t =  \sqrt{\frac{\sum_{i=0}^{W-1}{({s_{t-i} - \mu_t})^2}}{W-1}}.
\end{equation}
Similarly to Eq.~\ref{eq:mu}, we compute the empirical mean for a moving window $W'$, deemed $\Tilde{\mu}_t$. By design $W' \ll W$; i.e., $W$ and $W'$ are long- and short-term intervals, respectively. 

The likelihood of anomaly at time $t$, deemed $L_t$, is  
\begin{equation}\label{eq:Likelihood}
    L_t = 1 - Q\left(\frac{\Tilde{\mu}_t - \mu_t}{{\sigma}_t}\right), L_t\in(0,1),
\end{equation}
where  $Q$ is a Gaussian tail probability \cite{karagiannidis2007improved}. 
For a user-defined threshold $\epsilon$, if $L_t \geq 1 - \epsilon$, an observation at time $t$ is classified as anomalous.

\subsection{Experimental Setup}
\subsubsection{Feature Preparation and Dataset Split}

\begin{figure}[tb]
    \centering
    \includegraphics[width=1.0\columnwidth]{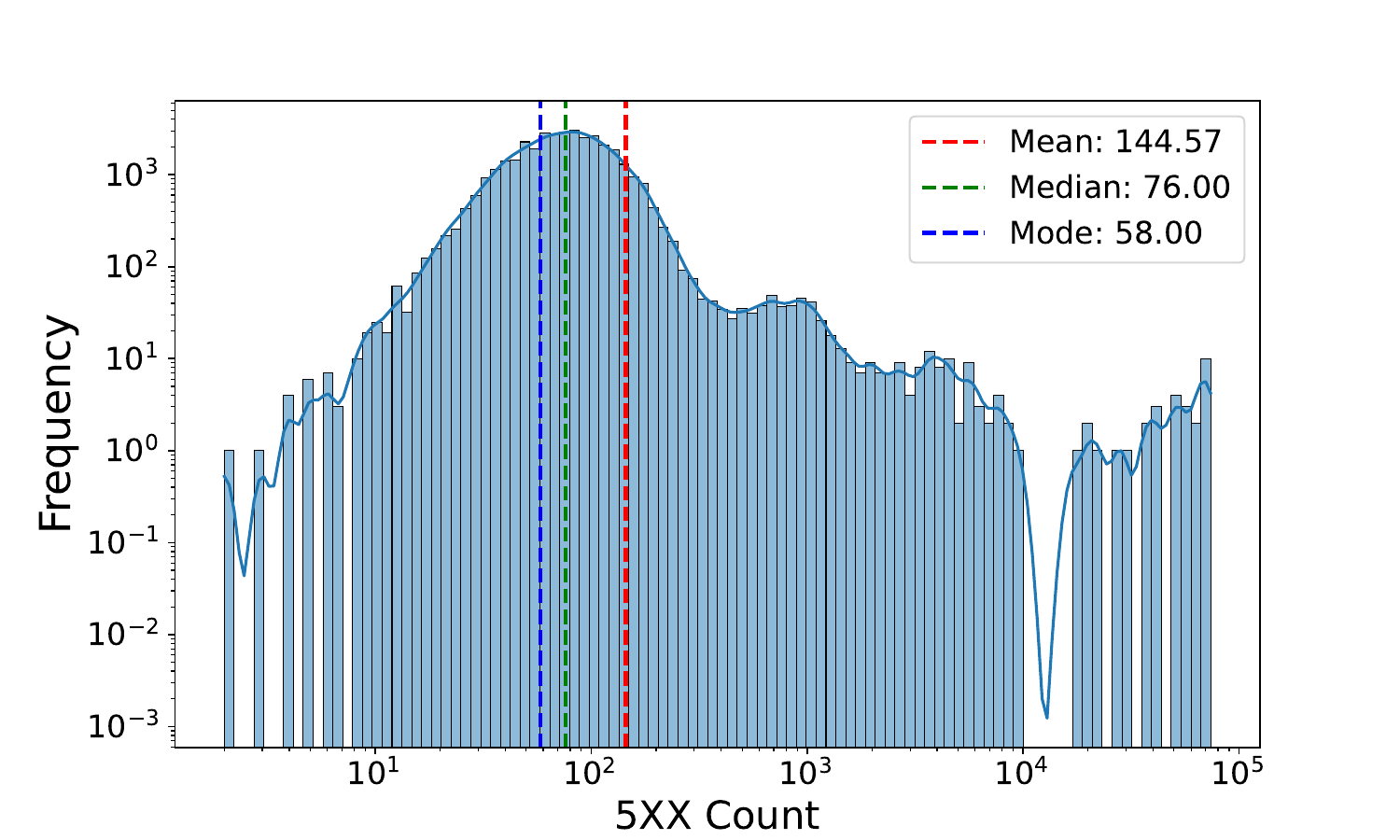}
    \caption{The distribution of request counts associated with 5XX errors. Both the $x$-axis and $y$-axis are logarithmically scaled, with the $x$-axis representing the number of 5XX errors and the $y$-axis showing the frequency of each corresponding error count. The solid blue line illustrates the overall distribution, highlighting the concentration and variation across different 5XX error values.}
    \label{fig:5xx_distribution}
\end{figure}

In the dataset features preparation, we kept only the subset of features corresponding to HTTP return codes in the 5XX range (server errors) and the ``count'' aggregate statistical function, resulting in a total of \num{2406} features. This step was done to reduce the dataset size based on the assumption that anomalous behavior is often reflected in changes in the frequency of server errors. To validate this assumption, we plot the distribution of the count of requests associated with 5XX codes in Figure~\ref{fig:5xx_distribution}. This distribution was obtained by summing all relevant features for each row, providing a cumulative error count for each timestamp. The right-skewed distribution, with a few timestamps exhibiting very high counts, suggests that anomalies are present in the data, making them a point of interest for anomaly detection purposes.

To capture weekly and daily periodicity, we added seasonality features using sine and cosine trigonometric functions~\cite{stolwijk1999studying}. This resulted in \num{2410} features in our input. The training data were scaled using min-max normalization, and the same normalization parameters were applied to the test data. Null values were replaced by zeros. The dataset was divided into five weeks of training data (from 2024-01-26 to 2024-02-29) and three months of test data (from 2024-03-01 to 2024-05-31). The test data contain 19 anomalies.

Each observation in the model represents data collected over an individual 5-minute interval.

\subsubsection{Evaluation Metrics}

We evaluated our models using the standard confusion matrix metrics: True Positive (TP), True Negative (TN), False Positive (FP), and False Negative (FN)~\cite{bishop_Springer_2006, o2013doing}. In this context, the positive label represents an anomaly, and the negative label denotes a non-anomaly.

However, in anomaly detection, relying solely on these point-based metrics is not ideal because detecting just one instance within an anomaly window is often sufficient in practice~\cite{ahmad2017unsupervised}. To address this, in addition to the conventional metrics, we also employed a window-based metric from the Numenta Anomaly Benchmark (NAB) to provide a more realistic evaluation of the models' performance. This metric is known as the \textit{NAB score}, ranging from 0 to 100, with higher numbers indicating better performance. Although not perfect~\cite{singh2017demystifying}, the NAB score is more practical for this task than metrics based on confusion matrices, such as accuracy or F1-score~\cite{bishop_Springer_2006}.

For a detailed explanation of the NAB score, refer to~\cite{ahmad2017unsupervised, lavin2015evaluating}. In brief, this metric rewards early detection (TP) within an anomaly window and penalizes false positives and false negatives. Any detection within an anomaly window is considered a TP, with its score determined by a scaled sigmoid function. A detection at the start of the anomaly window is assigned a score of 1, while detections near the end receive lower scores, closer to 0. Any subsequent detections within the same window are ignored for scoring purposes, meaning that all detections within one window count as a single TP. If the model misses an entire anomaly window, it results in one FN. FP and TN, on the other hand, are evaluated point-by-point. TNs do not affect the NAB score, but each FP lowers the score. 

The NAB score can be computed using different cost profiles; we use two shown in Table~\ref{tab:nab_profiles}: ``Standard'' and ``Reward Low FN''. The ``Standard'' profile strikes a balance between TP,  FN, and FP, applying a modest penalty on FP to prevent the model from favoring either precision or recall too heavily. In contrast, the ``Reward Low FN'' profile applies a harsher penalty on FN, which is useful in scenarios like ours where detecting each anomaly is crucial due to the importance of anomalies present in the dataset. This higher penalty on false negatives encourages the model to prioritize finding the anomalies, even at the risk of increasing FP count.

\begin{table}[tb]
    \caption{NAB score cost profiles as per~\cite{lavin2015evaluating}.}
    \label{tab:nab_profiles}
    \centering
    \begin{tabular}{@{}lrrrr@{}}
        \toprule
        Profile & TP Weight & FN Weight & FP Weight & TN Weight \\ \midrule
        Standard & 1.00 & 1.00 & 0.11 & 1.00 \\ 
        Reward Low FN & 1.00 & 2.00 & 0.11 & 1.00 \\ \bottomrule
    \end{tabular}
\end{table}

\subsubsection{Training and Testing Process}

We  trained the autoencoder on the training data. For the anomaly likelihood function, we set $L_t = 0.9996$, $W = 30$, and $W' = 2$; these values were chosen based on our prior experience with this type of task\footnote{Specifically, we conducted a small grid search using the following hyperparameter values: $L_t \in \{0.9990, 0.9995, 0.9996, 0.9997, 0.9998\}$, $W \in \{20, 25, 30, 35, 40, 50\}$, and $W' \in \{1, 2, 3, 4, 5\}$. While these represent a sample of potential model configurations, further performance improvements could be achieved with more extensive hyperparameter tuning. The chosen models and parameters are intended as illustrative examples, rather than as optimal configurations.}. The trained model was then tested on the test dataset.

\subsection{Results}

\begin{table*}[t]
    \caption{Performance of the anomaly detectors. AE denotes autoencoder.}
    \label{tab:static_setup_results}
    \centering
    \resizebox{\textwidth}{!}{
    \begin{threeparttable}
    \begin{tabular}{@{}lrrrrrrrrrr@{}}
        \midrule
        & \multicolumn{4}{c}{Confusion Matrix} & \multicolumn{2}{c}{NAB Score} &\multicolumn{3}{c}{Count of Detected Anomalies} \\ \cmidrule(lr){2-5} \cmidrule(lr){6-7} \cmidrule(l){8-10}
        Model & TP & TN & FP & FN & Standard Profile & Low FN Profile & Issue Tracker  & Instant Messenger & Test Log \\ 
        \midrule

                GRU AE & \num{8} & \num{25450} & \num{71} & \num{959} & \num{6.06} & \num{14.57} & 1 out of 3 & 5  out of 9 & 0  out of 7 \\ 
        ANN AE & \num{8} & \num{25445} & \num{76} & \num{959} & \num{4.61} & \num{13.60} & 1  out of 3 & 5  out of 9 & 0 out of 7 \\  

        \bottomrule
    \end{tabular}
	\end{threeparttable}

    }
\end{table*}

\subsubsection{Quantitative Results}

Table~\ref{tab:static_setup_results} presents the performance of the models in detecting anomalies.

The GRU model showed a slight improvement over the ANN model. The NAB score increased from 4.61 to 6.06 for the ``Standard Profile'' and from 4.61 to 13.60 for the ``Low FN Profile.'' However, the marginal improvements observed with the GRU model were somewhat unexpected, suggesting potential for further enhancement through hyperparameter tuning. Techniques such as advanced optimization strategies or refinements to the model architecture could better harness the GRU's capabilities and address current limitations in the experimental setup.

Both models detected 1 out of 3 customer-impacting anomalies from the Issue Tracker, 5 out of 9 anomalies from the Instant Messenger threads, and none of the 7 anomalies from the Test Log.

\begin{figure*}[t]
    \centering
    \includegraphics[width=\textwidth]{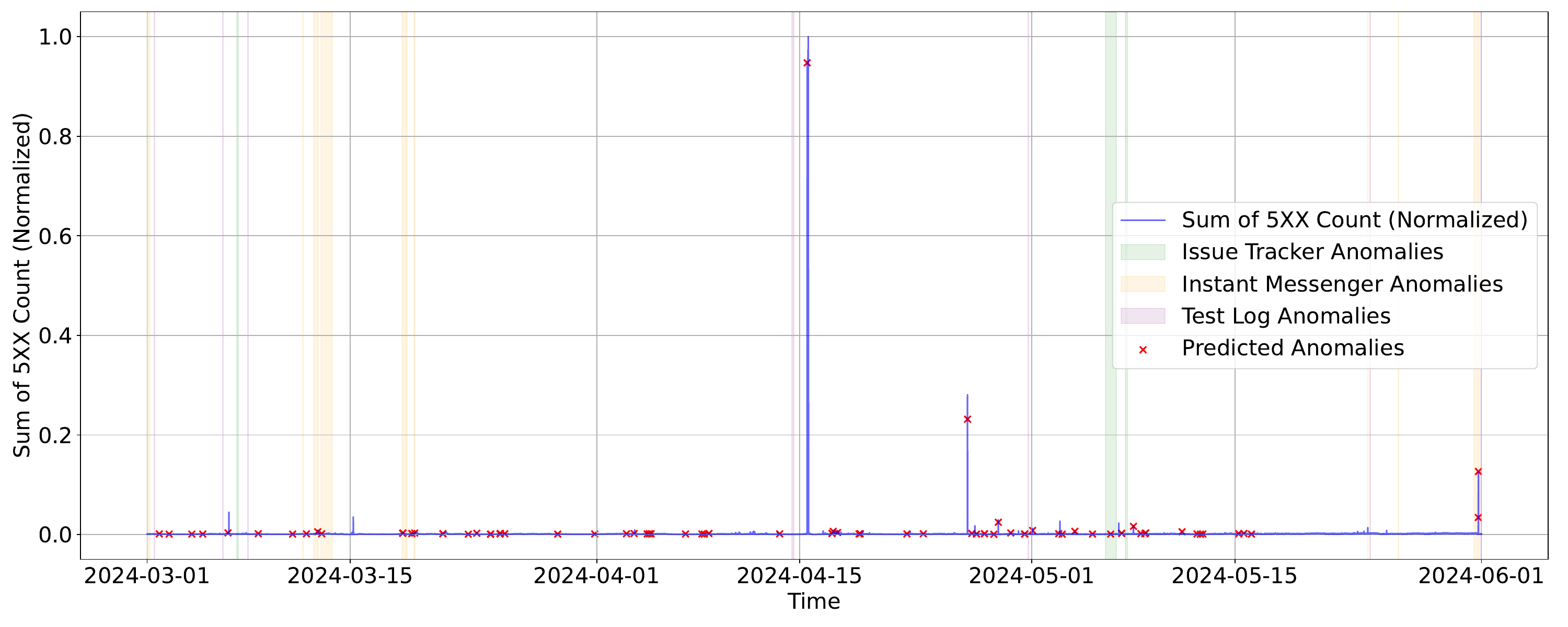}
    \caption{An example of anomaly detection using the GRU autoencoder. The $x$-axis represents the observation time. The $y$-axis shows (for simplicity of comprehension) the normalized sum of 5XX return codes, depicted by the blue line. Detected anomalies are marked with red crosses, while true anomaly windows are indicated by vertical bars in different colors, corresponding to the different sources of the anomaly reports.}
    \label{fig:anomaly_detection_results}
\end{figure*}

Figure~\ref{fig:anomaly_detection_results} illustrates an example of anomalies captured by the GRU autoencoder. The figure highlights both the anomalies the model successfully detects and those it misses. It also points out certain spikes in the number of 5XX errors that, despite being obviously abnormal, are not flagged as anomalies. There are several reasons for this. For example, some anomalies may not directly affect the \sus{}, while others are mitigated automatically through techniques such as high availability or caching.

\subsubsection{Qualitative Analysis}

We will highlight two example cases observed in the test data. The first case involves a  true operational issue successfully detected by the model. The second case, although flagged as anomalous by the model, shows unusual system behavior that does not correspond to any known issues in the ground truth data.

\begin{figure*}[tb]
    \centering
    \includegraphics[width=\textwidth]{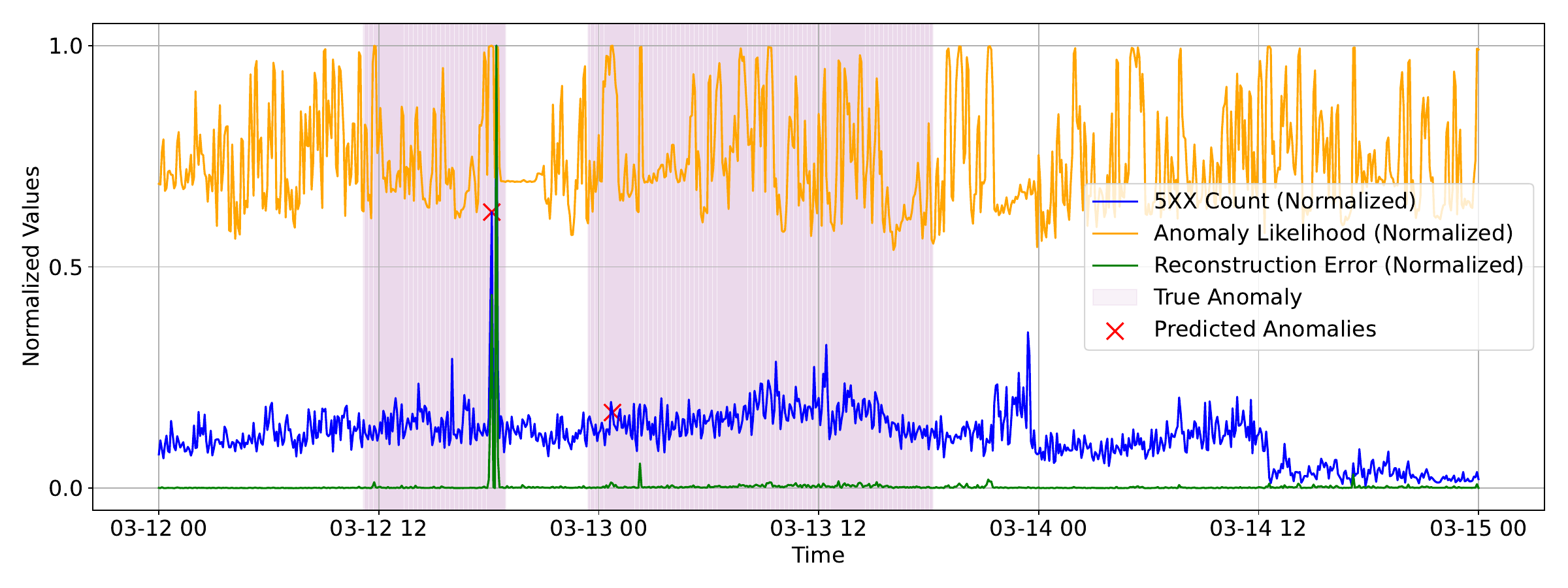}
    \caption{Example of two subsequent anomaly detections (present in the ground truth file) on March 12-13, 2024, using the GRU autoencoder model. The detected anomalies are highlighted with red crosses, while true anomaly windows are marked with purple bars. The blue line represents the normalized sum of the 5XX return codes count, the green line shows the reconstruction error, and the orange line indicates anomaly likelihood.}
    \label{fig:anomaly_detection_case1}
\end{figure*}

\paragraph{Case 1 (True Positives)}  \label{sec:tp_case}
During the period of March~12-13, 2024 (Figure~\ref{fig:anomaly_detection_case1}), the GRU-based anomaly detection model successfully flagged two anomaly windows. Both required taking specific instances of \sus{} offline and redirecting traffic to healthy instances. The model not only detected these extended anomaly windows but also identified the second anomaly early, demonstrating the model's ability to proactively catch issues before they escalate.

\begin{figure*}[tb]
    \centering
    \includegraphics[width=\textwidth]{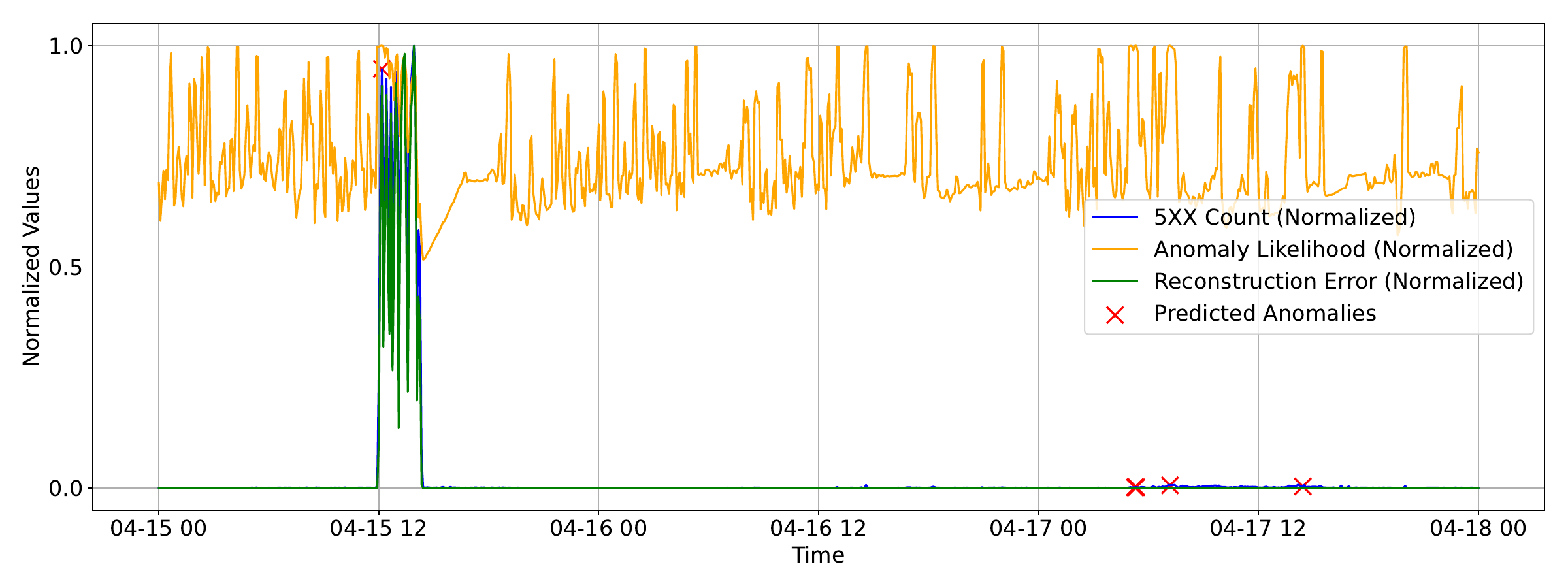}
    \caption{Example of detected abnormal behavior (absent from the ground truth file) from April~15-18, 2024, using the GRU autoencoder model. The blue line represents the normalized sum of the 5XX return codes count, the green line shows the reconstruction error, and the orange line indicates anomaly likelihood.}
    \label{fig:anomaly_detection_case2}
\end{figure*}

\paragraph{Case 2 (False Positives)}  \label{sec:fp_case}
Figure~\ref{fig:anomaly_detection_case2}  illustrates abnormal behavior detected on April 15-18, 2024, where the 5XX error count spiked to the highest level observed during the study. The spike lasted for nearly two hours and was accompanied by a sharp rise in reconstruction error, suggesting a significant deviation from normal patterns. The anomaly likelihood remained elevated throughout the event, which led the GRU model to flag it as anomalous.

While both cases present clear signs of abnormality (high error spikes and extended durations), only the first case corresponds to an actual operational issue.

Overall, these case examples highlight the challenges of modeling LCS behavior. Although the GRU model successfully identified unusual behavior, it underscores the need for further refinement to distinguish between true operational issues and abnormal patterns that do not require human intervention. Incorporating human feedback, for example, through a human-in-the-loop training model~\cite{islam2021anomaly,hrusto2022optimization,hrusto2023towards,hrusto2024autonomous}, could be essential in reducing false positives and improving the model’s reliability in real-world applications.

\subsection{Interpretation of Results}

A test of commonly used anomaly detection models shows low performance, with only 6 anomalies detected out of 19. Additionally, around 70 abnormalities were identified in the system (similar to those described in Section~\ref{sec:fp_case}), but these may not be useful to the IBM Ops team, as they could be resolved by various automatic mechanisms. \textit{This highlights the challenge of accurately detecting anomalies in the LCS and calls on the community to develop more sophisticated models to address this issue.}

\section{Insights and Challenges}\label{sec:challenges}

Our \textbf{RQ1} was: ``What are the key characteristics of telemetry datasets collected from LCS?’’ As discussed earlier, our dataset exhibits high dimensionality and volatility, reflecting the dynamic nature of LCS~---~complex, evolving systems with fluctuating workloads. These characteristics present significant challenges for anomaly prediction, leading us to  \textbf{RQ2}: “What are the main challenges in predicting anomalies within such large datasets?” Below, we explore some of these challenges and the corresponding insights.

The size of the system and the corresponding large feature set (on the order of $10^5$) make model training computationally expensive. This, in turn, complicates model hyperparameter tuning due to the high resource demands.  For example, during our experiments on the complete dataset, hyperparameter tuning for the trained models demanded significant computational resources, often resulting in prolonged training times and system bottlenecks, such as memory exhaustion or CPU overload during the deployment of anomaly detection systems. One possible solution is to reduce the dimensionality, either by selecting salient features based on expert knowledge or by employing automatic dimensionality reduction techniques. Developing scalable algorithms that balance computational efficiency and model performance remains a critical area for future work.

The IBM Cloud system we study is inherently non-stationary: the hardware and software stacks are constantly evolving, new features are regularly introduced, and the number of active users and their workloads fluctuate over time. Given the complexity of this system, unusual activities and failures often occur. However, the ground-truth data may not always reflect these activities and failures (see Section~\ref{sec:fp_case}).

As a result, anomaly detectors may flag issues that appear to be false positives when compared to ground-truth but are, in fact, genuine deviations from normal behavior.  An open challenge is determining how to automatically distinguish between two types of anomalies: those that impact customers versus those that are handled and resolved automatically\footnote{The latter type of anomalies~---~those that are automatically mitigated~---~are typically lower priority. However, they may still signal areas for potential system refactoring to improve robustness. This presents a common requirements prioritization dilemma: Should resources be allocated toward the development of new features or to reducing the number of failover events that are automatically mitigated and do not impact customers?}. Being able to make this distinction would reduce the burden on operations teams, particularly when triaging alerts during off-hours, which can be exhausting.

An alternative potential solution to address non-stationary is to incorporate data from multiple 5-minute intervals into each observation to capture more complex patterns and regularly retrain the models to address non-stationarity (similar to the approach in~\cite{islam2021anomaly}). However, this poses a challenge due to the high computational costs involved.

Another challenge is accurately identifying the start and end times of anomalies in such a complex system (as discussed in Section~\ref{sec:anomaly_annotation}). This is a known issue in real-world datasets (especially those having a high number of features)~\cite{lavin2015evaluating}. Although we attempt to identify the time windows of the anomalies, they are not always perfectly precise. This imprecision makes it harder for models to distinguish between normal and abnormal behavior. One approach is to introduce a ``time buffer'' around anomalous windows to prevent abnormal behavior from slipping into the training dataset of autoencoders. Metrics such as the NAB score help mitigate this to some extent, as reporting false positives near the end of a previous anomaly window results in a lower penalty.

\section{Threats to Validity}

This study, while providing valuable insights into anomaly detection within large-scale cloud systems, faces several potential threats to validity that should be acknowledged. We classify validity threats according to~\cite{wohlin2012experimentation,yin2009case}.

\subsection{External Validity}

Software engineering studies are often challenged by the variability of real-world environments, and the problem of \textit{generalization} cannot be fully resolved~\cite{wieringa2015six}. In this work, we consider our software and its associated telemetry dataset as a critical case\footnote{In case study research, a critical case refers to a particularly significant or decisive instance chosen for its potential to offer deep insights into a specific phenomenon~\cite{yin2009case}.}, which could assist the community in building anomaly detection models and testing them on the data from a real-world LCS. The same empirical evaluation and analysis methods can be applied to other software products, provided that well-designed and controlled experiments are conducted.

Beyond addressing the generalization problem, this dataset holds practical value for various stakeholders in cloud environments. For instance, if a client operates a complex cloud application (e.g., distributed and comprising multiple components), this dataset would remain relevant and of interest. Models trained at the provider level can help customers by identifying patterns or anomalies specific to distributed workloads, thereby improving the robustness and reliability of their operations. This highlights the practical value of the dataset beyond cloud providers, extending its applicability to end-users managing sophisticated cloud-based applications.

\subsection{Internal Validity}
Our dataset, while extensive, covers a limited time span of 4.5 months and represents telemetry data from a specific subset of IBM Cloud services. As such, it may not capture longer-term trends, rare failure modes, or seasonal variations that could influence the behavior of large-scale cloud systems. 

The trace data was system-generated, incorporating traces from both real users and synthetic test runs. However, the synthetic test cases used for generating telemetry data may not fully represent the complexity of real-world user interactions and may overlooks some issues. Thus, the synthetic test runs (which resulted in 11 ``Test Log'' anomalies) could bias our anomaly detection results, potentially leading to an over-representation or under-representation of certain types of incidents that typically occur in live environments. 

Despite these limitations, the 4.5-month time span remains valuable for identifying short-term patterns and providing insights into common issues.

\subsection{Construct Validity}
A challenge in this study is the difficulty of accurately defining ground truth for anomaly detection in LCS. 
For example, the actual start of an issue might be imprecise due to the reliance on thresholds or operational heuristics, and the termination of incidents might overshoot the real closing time. Additionally, some abnormal behavior may not affect end-users (e.g., due to high-availability mechanisms), further complicating the identification of true incidents. Despite these complexities, they reflect the realities of working with real-world datasets collected from complex systems.

\section{Conclusion}\label{sec:conclusion}

In this paper, we introduce a novel, large-scale dataset of real-world telemetry from IBM Cloud's Console software, a continuously evolving LCS. Collected over 4.5 months, the dataset captures aggregated response-time telemetry from microservices across multiple data centers. We intend for this dataset to serve as a new benchmark challenge for researchers and practitioners developing anomaly detection methods.

Our experiments utilized two predictive models for anomaly detection~---~ANN-based and GRU-based autoencoders. While both showed potential, they also highlighted key challenges, including high data dimensionality, non-stationary behavior, and difficulty in distinguishing between significant and insignificant anomalies in cloud systems.

More advanced techniques are required to effectively detect and predict anomalies that affect customer experience and system stability. Future research could investigate more sophisticated machine learning models, dimensionality reduction strategies, and active learning approaches to better manage the dynamic nature of cloud environments. Building a reliable anomaly detector that accurately predicts impactful anomalies remains an open research question, and we encourage the research community to contribute to this effort.

This study adds to the growing research on cloud anomaly detection by providing a benchmark dataset and addressing the practical difficulties of detecting anomalies in real-world cloud systems. We hope our dataset and findings will inspire the creation of more robust anomaly detection solutions, ultimately improving the reliability and performance of cloud services.

\bibliographystyle{IEEEtran}
\bibliography{References}

\appendices

\section{Data Preprocessing}\label{sec:data_preprocessing}
The collected data undergoes the following ``mutations'': 
\begin{enumerate*}[label=(\arabic*)]
    \item filtering,
    \item aggregation, 
    \item transformation, and
    \item masking. 
\end{enumerate*}
Each step is outlined below.

\subsection{Data Filtering}
The IBM Cloud \sus{} logs and telemetry traffic are monitored, and the relevant telemetry records are filtered based on criteria defined by IBM DevOps teams. The filtered data are stored in compressed gzip~\cite{gailly1992gnu} archives in COS. This reduces data points from thousands to approximately 200 per minute per data center, optimizing storage and processing needs. 

\subsection{Data Aggregation}\label{sec:data_aggr}
 The filtered data are aggregated into 5-minute intervals, aligned with astronomical time. Each interval (e.g., 0--5 minutes) groups incoming requests by rounding them to the start of the time range. The choice of a 5-minute interval is based on feedback from the IBM Ops team, as this duration smooths out noise and enhances the data’s usability. Eight aggregation functions~---~minimum, maximum, median, average, count, standard deviation, skewness, and kurtosis~---~are applied to prepare the data, specifically the response time (measured in milliseconds), for further analysis. 

\begin{figure*}[!t]
    \small
    \fbox{
    \begin{minipage}{1.0\textwidth}
    Template: \texttt{\{location\}\_\{kind\}\_\{host\}\_\{method\}\_\{statusCode\}\_\{endpoint\}\_\{aggregated\_stats\_name\}} \\
    Example: \texttt{datacenter1\_CLIENT\_component10\_GET\_200\_endpoint865\_count}
    \end{minipage}
    }
    \caption{Pivot dataset column template and column name example.}
    \label{fig:pivot_column_template}
\end{figure*}
\subsection{Data Transformation}
The telemetry data, originally in JSON format, is ingested into DuckDB~\cite{raasveldt2019duckdb}, a database management system optimized for efficient data analysis. Aggregated statistics, such as mean and median, are extracted from the JSON and then unpivoted into database columns. We share the resulting dataset in Apache Parquet format~\cite{apache_parquet} (due to its efficiency in storing and accessing large datasets).

\subsection{Dataset Masking}
To ensure privacy and comply with IBM's security policies, sensitive fields such as location, host, and endpoint (see Appendix~\ref{sec:aggr_telemetry_files} for details of the fields) are masked. These are replaced with obfuscated values like \texttt{datacenter1}, \texttt{component5}, and \texttt{endpoint8}.

\section{Dataset Characteristics}\label{sec:data_characteristics}

\subsection{Aggregated telemetry}\label{sec:aggr_telemetry_files}
The resulting dataset contains aggregated telemetry for \num{39365} 5-minute intervals. The data is provided in an unpivoted format in the file \texttt{unpivoted\_data.parquet}, which contains \num{413241248} rows and \num{9} columns:

\begin{itemize}
    \item \textbf{interval\_start}: The start time of a 5-minute interval, represented in Epoch/Unix time format.
    \item \textbf{location}: \num{7} distinct values, representing data center ID.
    \item \textbf{kind}: \num{2} distinct values, namely \texttt{CLIENT} and \texttt{SERVER}, corresponding to the communication type.
    \item \textbf{host}: \num{54} distinct host IDs.
    \item \textbf{method}: \num{7} distinct REST API~\cite{fielding2000rest,rfc7231} methods (e.g., \texttt{GET} or \texttt{POST}).
    \item \textbf{statusCode}: \num{30} distinct HTTP response status codes~\cite{rfc7231} (e.g., \texttt{200} or \texttt{500}) and \num{1} non-HTTP response status code (namely, \texttt{-1}). The count of columns for different groups of HTTP status codes in the pivoted version of the dataset is given in Table~\ref{tab:http_status_columns}.
    \item \textbf{endpoint}: \num{1001} distinct API endpoint IDs.
    \item \textbf{aggregated\_stats\_name}: One of eight aggregate statistical functions (detailed in Section~\ref{sec:dataset_description}).
    \item \textbf{aggregated\_stats\_value}: Contains the values corresponding to the aggregate statistical functions.
\end{itemize}

We also provide a pivoted version of the dataset in the file \texttt{pivoted\_data.parquet}, which contains \num{39365} rows and \num{117449} columns. Each row represents a specific time interval.

The first column is \texttt{interval\_start}, and the remaining column names are following the template shown in Figure~\ref{fig:pivot_column_template}. The values in these columns represent the corresponding \texttt{aggregated\_stats\_value}. 

Approximately 91\% of the cells in the pivoted dataset contain null values, due to inactivity or low activity\footnote{For example, calculating the standard deviation requires at least two data points, so if there are fewer, the value is null.} for a particular combination of location, kind, host, method, statusCode, endpoint, and aggregated statistical function.

\begin{table}[tb]
    \caption{Number of columns for different groups of HTTP status codes in the pivoted version of the dataset.}
    \label{tab:http_status_columns}
    \centering
    \begin{tabular}{@{}rrrrr@{}}
        \toprule
        2XX & 3XX & 4XX & 5XX & -1 \\ 
        Successful & Redirection & Client Errors & Server Errors & Non-HTTP \\
        \midrule
         \num{52288} & \num{4104} & \num{32680} & \num{19248} & \num{9128} \\ \bottomrule
    \end{tabular}
\end{table}

\subsection{Anomaly Windows}\label{sec:anomaly_windows_file}
Details of annotating anomaly windows are given in Section~\ref{sec:anomaly_annotation}.
The file \texttt{anomaly\_windows.csv} contains ground truth data, listing the time intervals when the \sus{} experienced anomalies. There are 25 anomalies in total. The file includes the following columns:
\begin{itemize}
    \item \textbf{number:} A unique identifier for each anomaly.
    \item \textbf{anomaly\_start:} The start time of the anomaly in ISO~8601 format~\cite{iso8601} (e.g., \texttt{2024-02-02 10:22:00-0500}).
    \item \textbf{anomaly\_end:} The end time of the anomaly in ISO~8601 format.
    \item \textbf{anomaly\_source:} The source of the anomaly. We assign numerical IDs to each source: 1 refers to ``Issue Tracker'', 2 refers to ``Instant Messenger'', and 3 refers to ``Test Log''; see Section~\ref{sec:anomaly_annotation} for details.
\end{itemize}

\subsection{The \sus{} Instances Downtime}

As mentioned in Section~\ref{sec:sus}, instances of the Console may be temporarily removed from or added to rotation in specific data centers. Note that removing an instance from rotation does not always indicate an anomaly; it could be a planned event, such as routine hardware or software maintenance. The file \texttt{location\_downtime.csv} provides details on when these events occur. In total, 93 events have been recorded in this file. This information can be useful for enhancing anomaly detection features or refining existing ones. The file includes the following columns:
\begin{itemize}
    \item \textbf{location}: The ID of the data center.
    \item \textbf{downtime\_start}: The start time of the downtime, in ISO~8601 format.
    \item \textbf{downtime\_end}: The end time of the downtime, in ISO~8601 format.
\end{itemize}

\end{document}